\documentclass[letterpaper]{article} 
\usepackage{aaai2026}  
\usepackage{times}  
\usepackage{helvet}  
\usepackage{courier}  
\usepackage[hyphens]{url}  
\usepackage{graphicx} 
\usepackage{natbib}  
\usepackage{caption} 
\usepackage{algorithm}
\usepackage{algorithmic}
\usepackage{marvosym}
\usepackage{amsmath}
\usepackage{tabularx}
\usepackage{booktabs}
\usepackage{multirow}
\usepackage{xcolor}
\usepackage{colortbl}
\usepackage{amssymb} 
\usepackage{pifont}
\usepackage{newfloat}
\usepackage{listings}
\DeclareCaptionStyle{ruled}{labelfont=normalfont,labelsep=colon,strut=off} 
\floatstyle{ruled}
\newfloat{listing}{tb}{lst}{}
\floatname{listing}{Listing}
\title{Point-SRA: Self-Representation Alignment for 3D Representation Learning}
\author{
    Lintong Wei\textsuperscript{\rm 1},~Jian Lu\textsuperscript{\rm 1}\thanks{Corresponding author.},~Haozhe Cheng\textsuperscript{\rm 2},~Jihua Zhu\textsuperscript{\rm 2},~Kaibing Zhang\textsuperscript{\rm 3}
}
\affiliations{
    \textsuperscript{\rm 1}School of Electronics and Information, Xi’an Polytechnic University\\
    \textsuperscript{\rm 2}School of Software, Xi’an Jiaotong University\\
    \textsuperscript{\rm 3}School of Computer Science, Xi’an Polytechnic University\\
    lujian@xpu.edu.cn
}

\begin{document}

\maketitle

\begin{abstract}
Masked autoencoders (MAE) have become a dominant paradigm in 3D representation learning, setting new performance benchmarks across various downstream tasks. Existing methods with fixed mask ratio neglect multi-level representational correlations and intrinsic geometric structures, while relying on point-wise reconstruction assumptions that conflict with the diversity of point cloud. To address these issues, we propose a 3D representation learning method, termed \textbf{Point-SRA}, which aligns representations through self-distillation and probabilistic modeling. Specifically, we assign different masking ratios to the MAE to capture complementary geometric and semantic information, while the \textbf{MeanFlow Transformer (MFT)} leverages cross-modal conditional embeddings to enable diverse probabilistic reconstruction. Our analysis further reveals that representations at different time steps in MFT also exhibit complementarity. Therefore, a \textbf{Dual Self-Representation Alignment} mechanism is proposed at both the MAE and MFT levels. Finally, we design a \textbf{Flow-Conditioned Fine-Tuning Architecture} to fully exploit the point cloud distribution learned via MeanFlow. Point-SRA outperforms Point-MAE by 5.37\% on ScanObjectNN. On intracranial aneurysm segmentation, it reaches 96.07\% mean IoU for arteries and 86.87\% for aneurysms. For 3D object detection, Point-SRA achieves 47.3\% AP@50, surpassing MaskPoint by 5.12\%.
\end{abstract}

\section{Introduction}

MAE has emerged as a leading framework in Self-Supervised Representation Learning (SSRL). In the 3D community, methods such as Point-MAE \cite{pang2022masked}, Point-M2AE \cite{zhang2022point}, and MaskPoint \cite{liu2022masked} have successfully adapted this paradigm and achieved strong performance across various tasks. By reconstructing masked regions from sparse visible points, 3D MAE learns robust geometric representations with strong generalization ability.
Despite its effectiveness, most existing methods adopt fixed masking ratio based on empirical settings, lacking theoretical insight into how the ratio influence the learned representations. This leads to an essential but underexplored question: \textit{Do different masking ratios inherently yield representations of varying richness, and can these differences be leveraged to improve overall representation quality?} 
\begin{figure}[t!]
  \centering
  \includegraphics[width=3.3in]{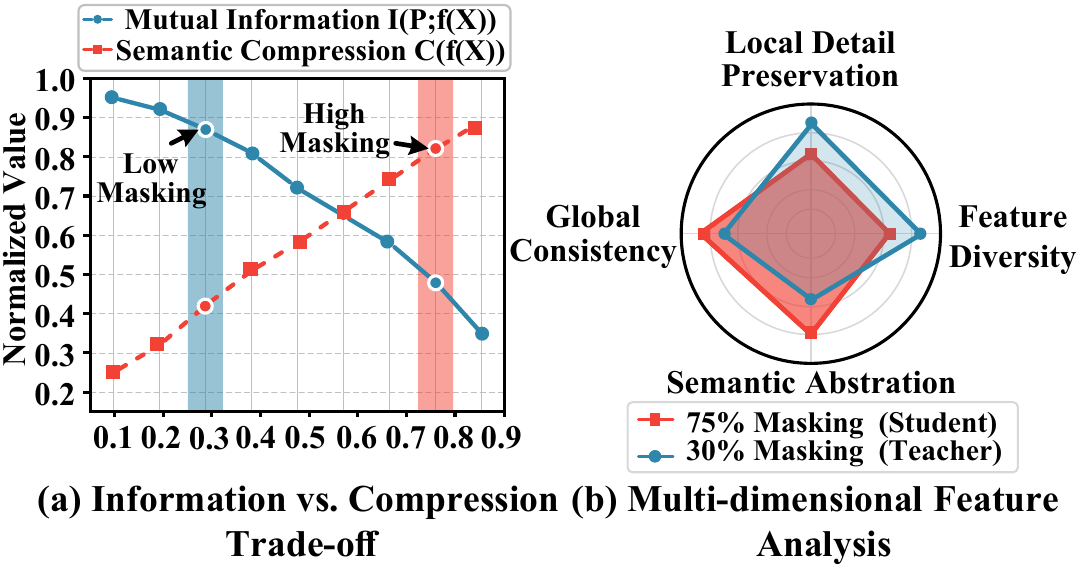}
 \caption{Complementarity under varying mask ratios. (a) Mutual information drops and semantic compression rises with increasing mask ratio. (b) 30\% mask (teacher) preserves fine-grained details; 75\% mask (student) yields better semantic abstraction.}
  \label{fig1}
\end{figure}

Our investigation reveals a fundamental principle: \textbf{masking ratio complementarity}. As shown in Figure \ref{fig1}, we discover that masking ratio choice creates a systematic trade-off between geometric detail preservation and semantic abstraction. Specifically, as masking ratio increases, mutual information with the input decreases while semantic compression improves, leading to representations that capture different aspects of 3D structure. Low masking ratios ($\leq$30\%) excel at preserving fine-grained geometric features, while high masking ratios ($\geq$75\%) force the model to learn abstract semantic patterns. This complementarity is not incidental but fundamental to the information bottleneck induced by masking. However, current 3D MAE methods suffer from two fundamental limitations that hinder their ability to utilize masking ratio complementarity effectively:

1. Fixed masking strategies limit the model’s ability to benefit from the complementary representations available under different masking ratios.

\begin{figure}[t!]
  \centering
  \includegraphics[width=3.2in]{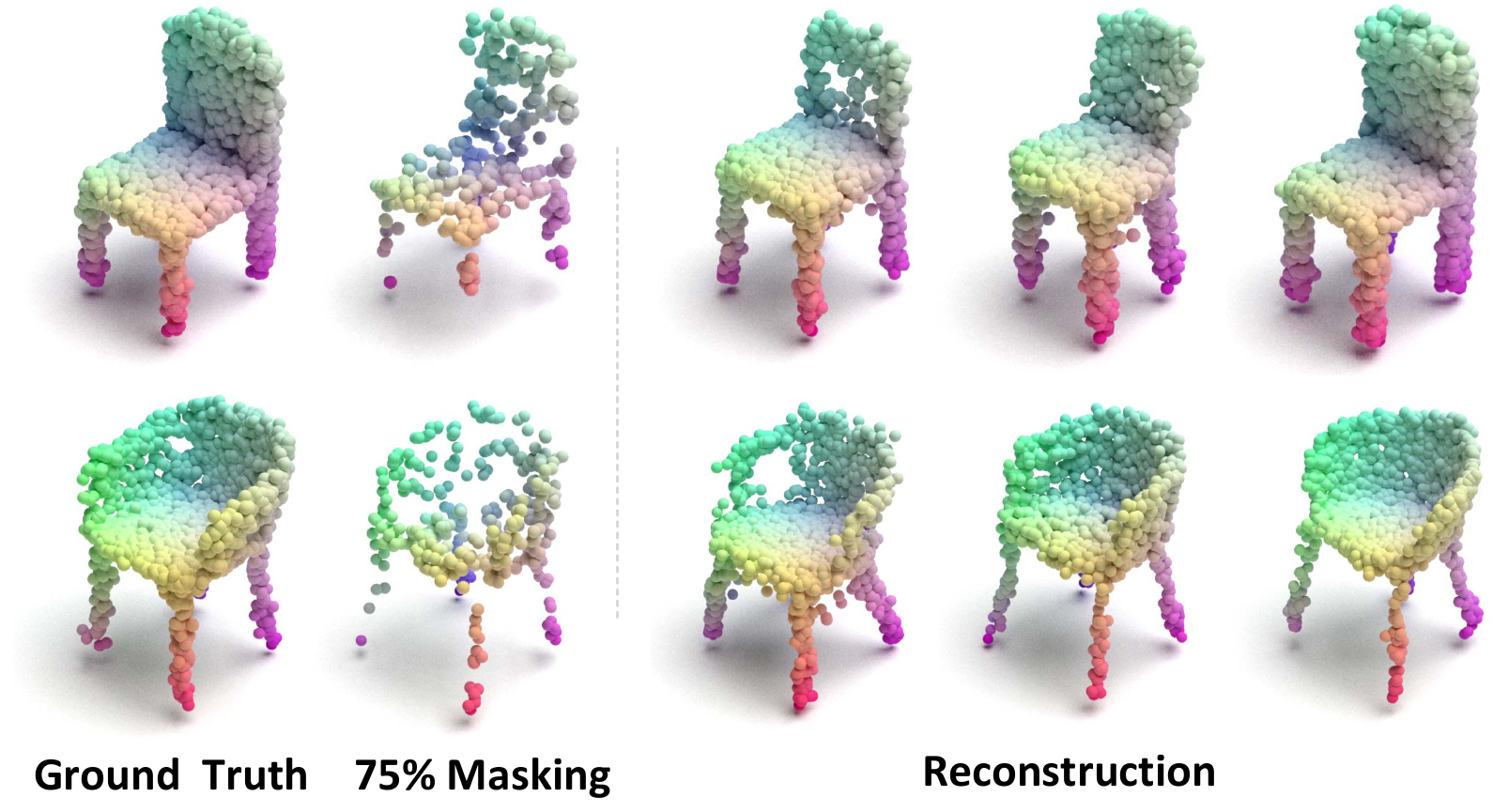}
 \caption{Under high masking ratios, diverse chair reconstructions from the same masked input (identical visible regions) reflect inherent uncertainty in geometric generation and plausible variations in leg shape, backrest angle, and seat thickness.}
  \label{fig2}
\end{figure}

2. Under high masking ratios, point-wise deterministic reconstruction in 3D MAE often leads to mismatches, as it fails to account for the inherent structural diversity. As illustrated in Figure \ref{fig2}, the same visible region may correspond to multiple plausible results.

To fully leverage the discovered masking ratio complementarity, we incorporate \textbf{Self-Representation Alignment (SRA)}  as an internal knowledge transfer mechanism. Meanwhile, we investigate the probabilistic reconstruction capabilities of MeanFlow and analyze representation information across different time steps. To this end, the \textbf{MeanFlow Transformer (MFT)} is designed to effectively leverage these properties. Moreover, MeanFlow probabilistic formulation naturally accommodates the multi-solution nature of 3D reconstruction. The trajectory-based learning process enables alignment of representations across temporal states, while its support for conditional generation allows flexible incorporation of multi-modal information. Therefore, we design a \textbf{Dual Self-Representation Alignment (Dual SRA)}  mechanism: \textbf{MAE-Level Self-Representation Alignment (MAE-SRA)} integrates representations obtained under different masking ratios, facilitating the fusion of geometric and semantic knowledge. \textbf{MFT-Level Self-Representation Alignment (MFT-SRA)} aligns representations across time steps using trajectories learned via MeanFlow, capturing the evolution of point cloud distributions.
Finally, we propose \textbf{Point-SRA}, a framework that systematically exploits masking ratio complementarity for 3D representation learning. In addition, the standard Transformer fine-tuning network is enhanced by integrating MeanFlow-based updates. Our main contributions are summarized as follows:
\begin{itemize}
\item We conduct a systematic theoretical analysis of masking ratio complementarity, reconstruction uncertainty, and numerical stability of MeanFlow, and leverage these insights to design the Point-SRA framework.
\item We present a unified Dual SRA mechanism for aligning representations across both masking ratios and temporal states, enabling fully self-contained knowledge transfer. 
\item We propose the MFT to address the limitations of point-wise reconstruction in 3D MAE, enabling representation guidance and probabilistic reconstruction.
\item We develop a Flow-Conditioned Fine-Tuning Architecture that leverages distributional knowledge learned during pre-training for downstream tasks.
\end{itemize}

\begin{figure*}[h!]
  \centering
  \includegraphics[width=6.8in]{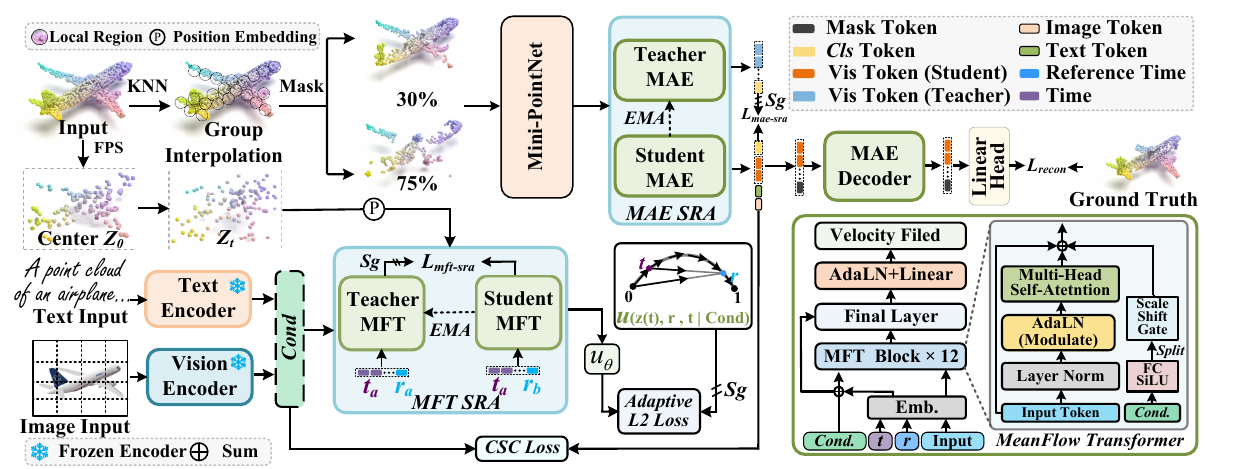}
 \caption{Overview of the Point-SRA. The input   point is partitioned via FPS and KNN. Image and text inputs are encoded by pre-trained Vision/Text Transformers. A dual self-representation alignment mechanism adopts a teacher-student MAE. The teacher uses a 30\% masking ratio to retain geometry, and the student uses 75\% to learn semantics. The teacher is updated via EMA. The MFT reconstructs probabilistic distributions and aligns representations across time. \textit{Sg} denotes gradient stop.}
  \label{fig3}
\end{figure*}
\section{Observations}
The point cloud $\mathcal{P}\in{\mathcal{R}}^{N\times3}$ is partitioned into $G$ local regions $\mathcal{N}_1,\mathcal{N}_2,...,\mathcal{N}_G$ via Farthest Point Sampling (FPS). A subset of $\lfloor r\cdot G\rfloor$ regions are randomly selected for masking, where the masking ratio $r\in[0,1]$. Let $\mathcal{V}$ denote the index set of visible regions, so the visible regions can be expressed as $\mathcal{X}_r=\{\mathcal{N}_i:i\in\mathcal{V}\}$. The encoder $f_{\theta_r}:\mathcal{X}_r\rightarrow\mathrm{R}^d$ maps the visible regions to a $d$-dimensional representation space.

\textbf{Theorem A: Masking Ratio Complementarity.} For masking ratios $r_l<r_h$, under the information bottleneck framework, the corresponding optimal encoders $f_{\theta_l^\ast}$ and $f_{\theta_h^\ast}$ satisfy:
\begin{equation}\mathcal{I}(\mathcal{P};f_{\theta_l^\ast}(\mathcal{X}_{r_l}))>\mathcal{I}(\mathcal{P};f_{\theta_h^\ast}(\mathcal{X}_{r_h})),\end{equation}
\begin{equation}\mathcal{C}(f_{\theta_h^\ast}(\mathcal{X}_{r_h}))>\mathcal{C}(f_{\theta_l^\ast}(\mathcal{X}_{r_l})),\end{equation}
where $\mathcal{I}(\cdot;\cdot)$ denotes mutual information and $\mathcal{C}(\cdot)$ denotes the semantic compression degree defined as:
\begin{equation}\mathcal{C}(Z)=\frac{\mathcal{H}(Z)}{\mathcal{I}(\mathcal{P}_{{semantic}};Z)},\end{equation}
where $\mathcal{P}_{\mathrm{semantic}}$ denotes semantic information of the point cloud. $\mathcal{H}(Z)$ is the entropy of representation $Z$. A detailed proof is provided by the Supplementary Material.

\textbf{Corollary: Representation Complementarity.} There exist projection functions: $\pi_{{geo}}:\mathcal{R}^d\rightarrow\mathcal{R}^{d_{{geo}}}$ and $\pi_{{sem}}:\mathcal{R}^d\rightarrow\mathcal{R}^{d_{{sem}}}$, such that:
\begin{align} 
    \| \pi_{{geo}}(f_{\theta_l^*}(\mathcal{X}_{rl})) & - \pi_{{geo}}(\mathcal{P}) \|_F \nonumber \\
    & < \| \pi_{{geo}}(f_{\theta_h^*}(\mathcal{X}_{rh})) - \pi_{{geo}}(\mathcal{P}) \|_F, 
\end{align}
\begin{align}
    \| \pi_{{sem}} & \! (f_{\theta_h^*}(\mathcal{X}_{rh})) - \pi_{{sem}}(\mathcal{P}) \|_F \nonumber \\
    & < \| \pi_{{sem}}\! (f_{\theta_l^*}(\mathcal{X}_{rl})) - \pi_{{sem}}(\mathcal{P}) \|_F,
\end{align} 
This provides the theoretical foundation for our Dual SRA mechanism, indicating that a high masking rate is conducive to preserving semantic information, while a low masking rate tends to retain geometric information.

\textbf{Theorem B: Reconstruction Uncertainty.} Traditional MAE methods are based on a point-wise reconstruction assumption, i.e., there exists a unique optimal reconstruction target:
\begin{equation}\mathcal{L}_{{det}}=E_{\mathcal{P}\sim p_{{data}}}\left[\parallel\mathcal{P}_{M}-G_{en}(\mathcal{P}_\mathcal{V})\parallel^2\right],\end{equation}
where $G_{en}$ is the generative function, and $\mathcal{P}_\mathcal{M}$ and $\mathcal{P}_\mathcal{V}$ denote the masked and visible regions, respectively. However, point cloud geometric reconstruction inherently exhibits ambiguity. Given visible regions $\mathcal{P}_\mathcal{V}$, the plausible configurations of the masked region form a conditional probability distribution:
\begin{equation}p(\mathcal{P}_\mathcal{M}|\mathcal{P}_\mathcal{V})=\int_{\Theta}p(\mathcal{P}_\mathcal{M}|\mathcal{P}_\mathcal{V},\omega)p(\omega|\mathcal{P}_\mathcal{V})d\omega,\end{equation}
where $\Theta$ denotes geometric parameters such as curvature and density. When the masking ratio lies within a reasonable range, the entropy of the reconstruction distribution satisfies:
\begin{equation}H(p(\mathcal{P}_\mathcal{M}|\mathcal{P}_\mathcal{V}))>0.\end{equation}
Naturally, we adopt MeanFlow to learn a continuous transformation from a noise distribution to the true data distribution, which can handle geometric uncertainty and generate diverse, geometrically consistent reconstructions.

\section{Methodology}
\subsection{Network Architecture}
As illustrated in Figure \ref{fig3}, the proposed Point-SRA framework consists of four tightly integrated components that work collaboratively to enhance 3D representation. The MAE serves as the foundational module, which learns geometric features by reconstructing the masked regions of the input. To address the inherent ambiguity in point cloud geometry, the MFT is designed to model the data distribution through continuous probabilistic trajectories, capturing geometric uncertainty more effectively. Building upon these two modules, the MAE-SRA aligns feature representations obtained under different masking ratios, enabling the fusion of geometric detail and semantic abstraction. Furthermore, the MFT-SRA aligns the probabilistic flow representations across temporal states, allowing the model to capture the dynamic evolution of the distribution.

\textbf{MAE.} A masking operation is first applied to randomly retain a subset of visible points $\mathcal{P}_{vis}=\{p_i|\mathcal{M}(p_i)=0\}$, while the rest are masked as $\mathcal{P}_{{mask}}=\{p_{i}|\mathcal{M}(p_{i})=1\}$. The encoder $E_{{MAE}}$ processes the visible points to extract their feature representations:
\begin{equation}h_{{vis}}=E_{{MAE}}(\mathcal{P}_{{vis}}).\end{equation}
At the same time, learnable mask tokens are generated for the masked points and concatenated with the visible features to form a complete feature set $h_{{full}}$. The decoder $D_{{MAE}}$ then takes this full representation as input to reconstruct the coordinates of all points $\hat{\mathcal{P}}$:
\begin{equation}\hat{\mathcal{P}}=D_{{MAE}}(h_{{full}}).\end{equation}
To improve reconstruction quality, the Chamfer Distance \cite{fan2017point} is adopted as the reconstruction loss:
\begin{equation}
\begin{split}
\mathcal{L}_{{recon}} = 
&\frac{1}{|\mathcal{P}|}\sum_{p_i\in\mathcal{P}} \min_{q_j\in\hat{\mathcal{P}}} \| p_i - q_j \|_2^2 \\
&+ \frac{1}{|\hat{\mathcal{P}}|}\sum_{q_j\in\hat{\mathcal{P}}} \min_{p_i\in\mathcal{P}} \| q_j - p_i \|_2^2
\end{split}
\end{equation}
where $p_i$ denotes a point from the ground-truth and $q_j$ denotes a point from the predicted point cloud.

\textbf{Conditional Distribution Modeling with MeanFlow.} Given a point cloud and its associated multi-modal conditional information, including image features $f_{{img}}$ and text features $f_{{text}}$, the conditional representations are first extracted using dedicated image and text encoders. A continuous trajectory  $\{z_t\}_{t\in[0,1]}$ is then defined. Where $z_0$ is the target point cloud and $z_1$ is sampled from a standard normal distribution. The trajectory is constructed via linear interpolation:
\begin{equation}z_t=(1-t)\cdot z_0+t\cdot z_1.\end{equation}

Subsequently, the MeanFlow method is employed to predict the average velocity field along this trajectory. Given a current time step $t$ and a reference time step $r$ with $r<t$, MeanFlow estimates the average velocity from $t$ to $r$ as:
\begin{equation}u_\theta\left(z_t,r,t\middle| c\right)\approx\frac{z_r-z_t}{r-t},\end{equation}
where $c$ denotes the conditional feature vector incorporating multi-modal information, constructed as follows:
\begin{equation}c=e_t\left(t\right)+e_r\left(r\right)+W_{\mathrm{img}}f_{{img}}+W_{{text}}f_{{text}},\end{equation}
where $e_t(\cdot)$ and $e_r(\cdot)$ are time embedding functions, and $W_{{img}}$,$\ W_{{text}}$ are projection matrices. The MFT is adopted as the backbone network to predict the mean velocity field, as illustrated in Figure \ref{fig3}.
\begin{equation}\mathcal{L}_{{MFM}}={E}_{t,r,z_t,z_r,c}\left[\parallel u_\theta\left(z_t,r,t\middle| c\right)-u_{{target}}\parallel^2\right].\end{equation}
The target velocity $u_{{target}}$ is derived from theoretical principles, not only considering the basic finite difference $\frac{z_r-z_t}{r-t}$, but also incorporating the time derivative of the instantaneous velocity field:
\begin{equation}u_{{target}}=v_t-(t-r)\cdot\frac{d}{dt}v_t(z_t),\end{equation}
where $v_t=z_1-z_0$, and the time derivative $\frac{d}{dt}v_t(z_t)$ is computed using the Jacobian-vector product (JVP):
\begin{equation}\frac{d}{dt}v_t\left(z_t\right)=\nabla_{z_t}v_t\left(z_t\right)\cdot v_t\left(z_t\right)+\frac{\partial v_t\left(z_t\right)}{\partial t}.\end{equation}
To stabilize training, we apply an Adaptive L2 loss, which dynamically adjusts the loss weight based on prediction error:
\begin{equation}\mathcal{L}_{{MFM}}={E}\left[{sg}\left(w\right)\cdot\parallel u_\theta-u_{{target}}\parallel^2\right],\end{equation}
where $w=\frac{1}{(\parallel u_\theta-u_{{target}}\parallel^2+\epsilon)^p}$, ${sg(}\cdot\mathrm{)}$ denotes the stop-gradient operation, $ p=1-\gamma$ is the power exponent, $\epsilon$ is a small constant for numerical stability, $u_\theta$ is the predicted average velocity field by MFT, and $u_{{target}}$ is the theoretical target computed via the MeanFlow identity.

\textbf{Cross-modal Joint Conditioning.} The geometric features of the point cloud are extracted via the MAE encoder, while image and text features are obtained as \textit{ImageTransformer} and \textit{TextTransformer}, respectively. These features are propagated across modules as follows:
\begin{equation}f_{{img-token}}=W_{{img}}(f_{{cls}}),\end{equation}
\begin{equation}f_{{text-token}}=W_{{text}}(f_{{cls}}),\end{equation}
where $f_{{cls}}$ denotes the $[CLS]$ token feature from the MAE encoder, and $W_{{img}},\ W_{{text}}$ are linear projection matrices. A Cross-modal Semantic Consistency (CSC) Loss is defined to align the point cloud encoder's representations with those from the image and text modalities:
\begin{equation}
\begin{aligned}
\mathcal{L}_{{CSC}} = 
&\mathcal{L}_{{SmoothL1}}(f_{{img}\text{-}{token}}, f_{{img}}) \\
&+ \mathcal{L}_{{SmoothL1}}(f_{{text}\text{-}{token}}, f_{{text}}).
\end{aligned}
\end{equation}

\textbf{Dual Self-Representation Alignment.} As illustrated in Figure \ref{fig3}, Self-Representation Alignment (SRA) is applied at both the MAE level and the MFT level to facilitate effective knowledge transfer. The teacher model parameters are updated from the student model via Exponential Moving Average (EMA):
\begin{equation}\theta_{{teacher}}\gets m\cdot\theta_{{teacher}}+\left(1-m\right)\cdot\theta_{{student}},\end{equation}
where $m\in[0,1]$ is a momentum coefficient controlling the update rate.

\textbf{MAE-SRA.} At the MAE level, a teacher-student architecture is built, where the student model is trained with a high masking ratio, and the teacher model with a low masking ratio. Formally, we apply two masking patterns $\mathcal{M}_{h}$ and $\mathcal{M}_{l}$ corresponding to high and low masking ratios, respectively, yielding masked inputs $\mathcal{P}_{h}$ and $\mathcal{P}_{l}$. The student model processes the high-mask input:
\begin{equation}h_{{student}}=E_{{MAE}}(\mathcal{P}_{h}).\end{equation}
The teacher model processes the low-mask input:
\begin{equation}h_{{teacher}}=E_{{MAE}}^{{EMA}}(\mathcal{P}_{l}),\end{equation}
where $E_{{MAE}}^{{EMA}}$ denotes the teacher encoder. The MAE-SRA loss is defined as the cosine similarity loss between the feature representations of the student and teacher models:
\begin{equation}\mathcal{L}_{{mae-sra}}=1-\frac{h_{{student}}\cdot h_{{teacher}}}{|h_{{student}}|\cdot|h_{{teacher}}|}.\end{equation}

\textbf{MFT-SRA.} At the MFT level, a temporal alignment strategy is designed to align representations at different time points and capture the evolution of the probabilistic flow. Specifically, we select two distinct time steps $t_a>t_b$, where the student model processes the representation at $t_a$, and the teacher model processes that at $t_b$.
Given the condition $c$, and the corresponding noisy point cloud states $z_{t_a}$and$ z_{t_b}$, the feature representations are extracted as:
\begin{equation}h_{t_a}=F_{{MF}}\left(z_{t_a},t_a,c\right),h_{t_b}=F_{{MF}}^{{EMA}}(z_{t_b},t_b,c),\end{equation}
where $ F_{{MF}}$ and $F_{{MF}}^{{EMA}}$ denote the feature extractors of the student and teacher MFT networks, respectively. The MFT-SRA loss is then defined as the cosine similarity loss between the two representations:
\begin{equation}\mathcal{L}_{mft-sra~}=\parallel h_{t_a}-{sg}(h_{t_b}+u_\theta(z_{t_b},t_a,t_b|c)\cdot(t_a-t_b))\parallel^2.\end{equation}

\textbf{Joint Loss Optimization.} The final training objective is formulated as a weighted sum of multiple loss components:
\begin{equation}
\begin{aligned}
\mathcal{L}_{{total}} &= 
\mathcal{L}_{{recon}} + \lambda_{{flow}}\mathcal{L}_{{MFM}} \\
&+ \mathcal{L}_{{CSC}}+ \lambda_{{mae\text{-}sra}}\mathcal{L}_{{mae\text{-}sra}}+ \lambda_{{mft\text{-}sra}}\mathcal{L}_{{mft\text{-}sra}},
\end{aligned}
\end{equation}
where $\lambda_{{flow}}$ is set to 0.5, and both $\lambda_{{mae-sra}}$ and $\lambda_{mft{-sra}}$ are set to 0.2.
\begin{figure}[t!]
  \centering
  \includegraphics[width=3.3in]{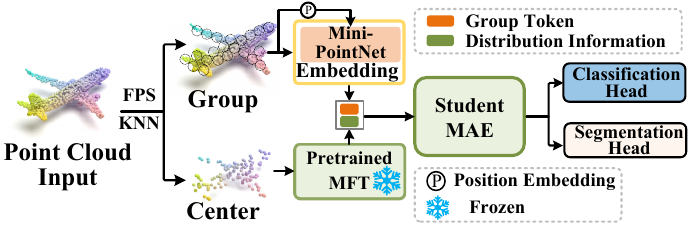}
 \caption{Flow-conditioned Fine-Tuning Architecture of Point-SRA. The pretrained MFT (with frozen parameters) takes the center coordinates and outputs flow vectors sampled at different time steps.}
  \label{fig4}
\end{figure}
\begin{table}[h!]
    \centering
    \small
    \renewcommand\arraystretch{1.0}
    \begin{tabular}[width=0.95\linewidth]{p{2.8cm} p{0.8cm}<{\centering}p{0.8cm}<{\centering}p{0.8cm}<{\centering}p{1.1cm}<{\centering}}
        \toprule
        \toprule
        \multirow{2}{*}{Method} & \multicolumn{3}{c}{ScanObjectNN} & \multirow{2}{*}{Params (M)}\\
        \cmidrule(lr){2-4}
         & BG & ONLY & RS & \\
        \midrule
            \multicolumn{5}{c}{\textit{Supervised Learning Only}} \\ 
            \midrule
            PointNet~\shortcite{qi2017pointnet} & 73.30 & 79.20 & 68.00 & 3.5 \\ 
            PointNet++~\shortcite{qi2017pointnet++} & 82.30 & 84.30 & 77.90 & 1.5 \\  
            PointNeXt~\shortcite{qian2022pointnext} & -- & -- & 87.7±0.4 & 1.4 \\ 
            P2P~\shortcite{wang2022p2p}  & -- & -- & 89.30 & 195.8 \\
            \midrule
            \multicolumn{5}{c}{\textit{Single-Modal Self-Supervised Representation Learning}} \\ 
            \midrule
            Point-BERT~\shortcite{yu2022point} & 87.43 & 88.12 & 83.07 & 22.1 \\
            Point-MAE~\shortcite{pang2022masked}  & 90.02 & 88.29 & 85.18 & 22.1 \\
            Point-M2AE~\shortcite{zhang2022point}  & 91.22 & 88.81 & 86.43 & 15.3 \\
            PointMamba~\shortcite{liang2024pointmamba} & 94.32 & 92.60 & 89.31 & 12.3 \\
            PointDif~\shortcite{zheng2024point}  & 93.29 & 91.91 & 87.61 & -- \\ 
            Point-FEMAE~\shortcite{zha2024towards}  & 95.18 & 93.29 & 90.22 & 41.5 \\ 
            Point-JEPA~\shortcite{saito2025point}& 93.20 & 91.90 & 87.60 & -- \\ 
            RI-MAE~\shortcite{su2025ri} & 91.90 & -- & -- & -- \\ 
           \midrule
            \multicolumn{5}{c}{\textit{Cross-Modal Self-Supervised Representation Learning}} \\ 
            \midrule
            ACT~\shortcite{dong2022autoencoders}  & 93.29 & 91.91 & 88.21 & 22.1 \\
            I2P-MAE\shortcite{zhang2023learning} & 94.15 & 91.57 & 90.11 & 15.3 \\
            ReCon~\shortcite{qi2023contrast}  & 95.18 & 93.29 & 90.63 & 44.3 \\
            \textbf{Point-SRA (Ours)}  & \textbf{95.53} & \textbf{93.31} & \textbf{90.77} & 40.1\\
            \bottomrule
        \bottomrule
    \end{tabular}
    \caption{Classification accuracy (\%) on three subsets of ScanObjectNN. “–” indicates that the result is not available.}
    \label{table:1}
\end{table}

\textbf{Flow-Conditioned Fine-Tuning Architecture.} As illustrated in Figure \ref{fig4}, during fine-tuning, The pre-trained MFT is employed to compute the flow vector for each point cloud group. Unlike the pre-training stage, fine-tuning utilizes only point cloud geometry and does not rely on image or text modalities:
\begin{equation}F_{u_\theta}={{MFT}}_{{frozen}}(Center,t,r),\end{equation}
where $Center\in\mathcal{R}^{G\times3}$ denotes the center coordinates of each point cloud group, and ${{MFT}}_{{frozen}}$ indicates the pre-trained MFT module with frozen parameters. The time parameters $t$ and $r$ are sampled as in pre-training. To project the 3D flow vectors into the feature space, we design a dedicated projection layer:
\begin{equation}{F}_{cond}={MLP}({F}_{u_\theta}+\|{F}_{u_\theta}\|_2),\end{equation}
where $\parallel F_{u_\theta}\parallel_2$ is the L2 norm of the flow vector applied element-wise or broadcast to match dimensions. In addition, an adaptive gating mechanism is introduced to control the influence of flow conditioning on the original group features:
\begin{equation}g=\sigma\left({{MLP}}_{{gate}}\left(F_{{cond}}\right)\right),\end{equation}
\begin{equation}H_\mathrm{e}=H_\mathrm{g}\odot(1+\alpha\ g)+\beta F_{{cond}},\end{equation}
where $H_\mathrm{g}$ is the original group feature, $g\in[0,1]^{G\times d}$ is the learned gating value, and $\alpha$, $\beta$ are trainable modulation parameters. The operator $\odot$ denotes element-wise modulation.

\section{Experiments}
\textbf{Object Classification.} The pre-trained model is evaluated on the real-world object classification benchmark. ScanObjectNN \cite{uy2019revisiting} comprises 15K point cloud objects across 15 categories and is divided into three subsets: OBJ\_BG (objects with background), OBJ\_ONLY (objects only), and PB\_T50\_RS (objects with background and human-induced perturbations). As shown in Table \ref{table:1}, our method achieves overall accuracies of 95.53\%, 93.31\%, and 90.77\% on OBJ\_BG, OBJ\_ONLY, and PB\_T50\_RS respectively, surpassing Point-MAE \cite{pang2022masked} by +5.51\%, +5.02\%, and +5.59\%. 

\begin{table}[t!]
    \small
    \centering
    \begin{tabular}{p{2.4cm}p{0.3cm}<{\centering}p{0.4cm}<{\centering}p{0.4cm}<{\centering}p{0.4cm}<{\centering}p{0.4cm}<{\centering}p{0.35cm}<{\centering}p{0.35cm}<{\centering}}
        \toprule
        \toprule
        \multirow{3}*{\centering Method} & \multicolumn{3}{c}{Classification} & \multicolumn{4}{c}{Segmentation} \\
        \cmidrule(lr){2-4} \cmidrule(lr){5-8}
        & \multirow{2}{*}{V\%} & \multirow{2}{*}{A\%} & \multirow{2}{*}{F1\%} & \multicolumn{2}{c}{IoU\%} & \multicolumn{2}{c}{DSC\%} \\
        \cmidrule(lr){5-6} \cmidrule(lr){7-8}
        & & & & V& A & V & A\\
        \midrule 
        PointNet\shortcite{qi2017pointnet} & 93.7 & 69.5 & 69.2 & 74.2 & 37.8 & 84.2 & 49.6 \\
        PointNet++\shortcite{qi2017pointnet++} & 98.8 & 87.3 & 90.2 & 93.2 & 76.2 & 96.4 & 84.6 \\
        SO-Net\shortcite{li2018so} & 98.9 & 83.9 & 88.5 & 94.5 & 81.4 & 97.1 & 88.8 \\
        PointCNN\shortcite{li2018pointcnn} & 99.0 & 85.8 & 90.4 & 93.6 & 73.6 & 96.6 & 81.4 \\
        Point-MAE\shortcite{pang2022masked} & 97.9 & 89.4 & 93.7 & 89.1 & 67.7 & 93.9 & 75.6 \\
        PointDif\shortcite{zheng2024point} & 97.0 & 63.6 & 83.1 & 75.2 & 33.7 & 85.1 & 43.5 \\
        \midrule
        ACT\shortcite{dong2022autoencoders} & 97.3 & 84.9 & 91.4 & 85.4 & 57.7 & 91.7 & 67.9 \\
        ReCon\shortcite{qi2023contrast} & 99.1 & 93.9 & 96.8 & 95.7 & 84.7 & 97.8 & 91.2 \\
        \textbf{Point-SRA } & \textbf{100} & \textbf{95.1} & \textbf{97.7} & \textbf{96.1} & \textbf{86.9} & \textbf{98.0} & \textbf{92.7} \\
        \bottomrule
        \bottomrule
    \end{tabular}
    \caption{Classification and segmentation results for vascular and aneurysm segments on the IntrA.}
    \label{intra}
\end{table}
\textbf{Intracranial Aneurysm Classification and Segmentation.} The IntrA \cite{yang2020intra} contains 1,909 vessel segments, including 1,694 healthy segments and 215 aneurysmal segments. The F1-Score (F1) is used as the primary evaluation metric. V and A represent the classification accuracy for healthy intracranial vessels and aneurysms, respectively. IoU and Dice Similarity Coefficient (DSC) are employed to assess segmentation performance for vessel and aneurysm regions. As shown in Table \ref{intra}, Point-SRA achieves an F1-Score of 97.7, outperforming the unimodal PointMAE by 4, and the multi-modal ACT \cite{dong2022autoencoders} method by 6.3. In the segmentation task, Point-SRA attains an IoU of 86.9\% and a DSC of 92.7\% for aneurysms, both significantly surpassing mainstream segmentation baselines.

\begin{table}[t]
    \setlength{\tabcolsep}{4pt}
    \setlength{\extrarowheight}{1pt}
    \centering
    \small
    	\begin{tabular}[width=0.80\linewidth]{p{3.0cm} p{1cm}<{\centering}p{1cm}<{\centering}p{1cm}<{\centering}p{1cm}<{\centering}}
		\toprule
           \toprule
          \multirow{2}{*}{Method}  & \multicolumn{2}{c}{5-way}& \multicolumn{2}{c}{10-way}\\
\cline{2-3} \cline{4-5}
              &10-shot&20-shot&10-shot&20-shot\\
\toprule
        Point-BERT\shortcite{yu2022point}&94.6±3.1&96.3±2.7&91.0±5.4&92.7±5.1\\
        MaskPoint\shortcite{liu2022masked}&95.0±3.7& 97.2±1.7&91.4±4.0&93.4±3.5\\
        Point-MAE\shortcite{pang2022masked}&96.3±2.5&97.8±1.8& 92.6±4.1&95.0±3.0\\
        Point-M2AE\shortcite{zhang2022point}&96.8±1.8&98.3±1.4& 92.3±4.5&95.0±3.0\\
        RMRL\shortcite{wang2024rethinking}&97.2±2.3&98.7±1.2&93.2±3.4&95.6±2.6\\
        PointDif\shortcite{zheng2024point}& 96.4±1.8 & 98.0±1.4 & 92.3±4.5 & 95.2±3.0 \\
        \midrule
         ACT\shortcite{dong2022autoencoders}& 96.8±2.3 & 98.0±1.4 & 93.3±4.0 & 95.6±2.8 \\
        ReCon\shortcite{qi2023contrast}& 97.3±1.9 & 98.9±1.2 & 93.3±3.9 & 95.8±3.0 \\
        \textbf{Point-SRA(Ours)} & \textbf{97.6±2.2} &\textbf{99.0±1.2} & \textbf{93.3±4.5} &\textbf{95.9±2.7} \\
       \toprule
        \toprule
    \end{tabular}
    \caption{ Few-shot classification results on ModelNet40 are presented. The average accuracy (\%) is reported under both 5-way and 10-way settings with 10-shot and 20-shot configurations.
}
    \label{table_few_shot}
\end{table}
\textbf{Few-Shot Classification.} Systematic experiments are conducted on ModelNet40 \cite{wu20153d} under few-shot learning scenarios. As shown in Table \ref{table_few_shot}, under the 5-way configuration, Point-SRA achieves an average accuracy of 97.6\% and 99.0\% in the 10-shot and 20-shot settings, respectively, outperforming the Point-MAE \cite{pang2022masked} baseline by 1.3\% and 1.2\%. In the 10-way configuration, Point-SRA reaches 93.3\% and 95.9\% accuracy in the 10-shot and 20-shot settings, respectively.

\begin{table}[t!]
\renewcommand\arraystretch{1.2}
    \small
	\begin{tabular}[width=0.80\linewidth]{p{2.8cm} p{2.3cm}<{\centering}p{2.0cm}<{\centering}}
		\toprule
           \toprule
          Method&Pre Dataset&ScanNet(mIoU\%)\\
\toprule
           VoteNet\shortcite{ding2019votenet}&-&35.5\\
           STRL\shortcite{huang2021spatio}&ScanNet&38.4\\
           PointContrast\shortcite{xie2020pointcontrast}&ScanNet&38.0\\
           DepthContrast\shortcite{zhang2021self}&ScanNet-vid&42.9\\
        \midrule
           Point-BERT\shortcite{yu2022point}&ScanNet-Medium&38.3\\
           MaskPoint\shortcite{liu2022masked}&ScanNet-Medium&42.1\\
           Point-MAE\shortcite{pang2022masked}&ShapeNet&42.8\\
           PointDif\shortcite{zheng2024point}&ShapeNet&43.7\\ 
           \textbf{Point-SRA(Ours)}&ShapeNet&\textbf{47.4}\\
\toprule
        \toprule
	\end{tabular}
    \caption{Object detection results (AP@50) on the ScanNetV2.}
        \label{object}
\end{table}
\textbf{3D Object Detection.} In the indoor scene detection task on the ScanNetV2 \cite{dai2017scannet}, a 3D sparse convolutional network based on MinkowskiEngine \cite{choy20194d} is adopted as the backbone, and the Point-SRA pre-trained encoder is integrated into the feature extraction pipeline. The ScanNetV2 consists of 1,513 fully scanned indoor scenes, covering 18 common object categories. We follow the standard Average Precision (AP) metric and evaluate under an IoU threshold of 0.5 (AP@50). Table \ref{object} show the Point-SRA pre-trained encoder achieves an AP@50 of 47.4\%. This improvement mainly comes from more accurate localization and recognition of objects with complex geometric structures, demonstrating the effectiveness of Point-SRA’s geometric feature representations in modeling spatial complexity.

\begin{table}[t!]
\renewcommand\arraystretch{1.2}
    \small
	\begin{tabular}[width=0.80\linewidth]{p{2.5cm} p{1.6cm}<{\centering}p{1cm}<{\centering}p{1.6cm}<{\centering}}
		\toprule
           \toprule
          Method&Pre-training& mIoU(\%)&mAcc(\%)\\
\toprule    
           PointNet\shortcite{qi2017pointnet}&\ding{55}&41.1&49.0\\
           
           PointCNN\shortcite{li2018pointcnn}&\ding{55}&57.3&63.9\\
           SegGCN\shortcite{lei2020seggcn}&\ding{55}&63.6&70.4\\
           Pix4Point\shortcite{qian2022pix4point}&\ding{55}&69.6	&75.2\\
           MKConv\shortcite{woo2023mkconv}&\ding{55}&67.7&75.1\\
           PointNeXt\shortcite{qian2022pointnext}&\ding{55}&68.5&75.1\\
        \midrule
           Point-BERT\shortcite{yu2022point}&$\checkmark$	&68.9	&76.1\\
           MaskPoint\shortcite{liu2022masked}&$\checkmark$ &68.6	&74.2\\
           Point-MAE\shortcite{pang2022masked}&$\checkmark$&	68.4&	76.2\\
           PointDif\shortcite{zheng2024point}&$\checkmark$&	70.0&	77.1\\
           \textbf{Point-SRA(Ours)}&$\checkmark$&\textbf{71.8}&\textbf{79.3}\\
\toprule
        \toprule
	\end{tabular}
    \caption{Semantic segmentation results on S3DIS Area-5. mIoU and mAcc represent mean Intersection over Union and mean accuracy, respectively.}
\label{semantic seg}
\end{table}
\textbf{Indoor Scene Segmentation.} The S3DIS \cite{armeni20163d} consists of 6 large indoor areas with a total of 271 rooms, covering 13 semantic categories and furniture objects. We follow the standard Area-5 evaluation protocol, using Area-5 as the test set and the remaining areas for training. In our experiments, a sparse convolutional network backbone based on MinkowskiEngine \cite{choy20194d} is adopted, keeping the same decoder architecture and segmentation head, while the encoder is replaced with the pre-trained Point-SRA model. During training, the Point-SRA encoder is frozen, and only the decoder and segmentation head are fine-tuned. The AdamW optimizer is used with an initial learning rate of 0.006 and weight decay of 0.05, and training is conducted for 100 epochs. As shown in Table \ref{semantic seg}, the results demonstrate that Point-SRA achieves 71.8\% mIoU and 79.3\% mAcc on the S3DIS test set.  

\begin{table}[t!]
\renewcommand\arraystretch{1.2}
    \small
	\begin{tabular}[width=0.70\linewidth]{p{2.1cm}p{1.3cm}<{\centering}p{1.6cm}<{\centering}p{1.6cm}<{\centering}}
		\toprule
           \toprule
           &OBJ\_BG&OBJ\_ONLY&PB\_T50\_RS\\
\toprule    
           Baseline &90.02&88.29&85.18\\
           MeanFlow &95.18&92.77&90.63\\
           MAE-SRA &95.01&92.77&89.69\\
           MFT-SRA & 95.35 & 92.91 & 90.01\\
           \textbf{Point-SRA} &\textbf{95.53} & \textbf{93.31} & \textbf{90.77}\\
\toprule
        \toprule
	\end{tabular}
    \caption{Ablation study results on core components.}
    \label{Submodule}
\end{table}
\subsection{Ablation Study}
\textbf{Effectiveness of Core Components.} Key components of Point-SRA are progressively removed to quantify each module’s contribution to the final performance. Point-MAE serves as the baseline, with all components utilizing the Flow-Conditioned Fine-Tuning Architecture. Table \ref{Submodule} presents the performance under different component combinations. MeanFlow brings a significant improvement, boosting accuracy on PB\_T50\_RS by 5.45\%, validating the effectiveness of probabilistic reconstruction over point-wise reconstruction. The complete Point-SRA framework achieves substantial gains across all tasks compared to the baseline MAE, demonstrating the effectiveness of component integration and the soundness of the overall design.

\begin{table}[t!]
\renewcommand\arraystretch{1.2}

    \small
	\begin{tabular}[width=0.70\linewidth]{p{2.2cm}p{0.7cm}<{\centering}p{1.8cm}<{\centering}p{1.9cm}<{\centering}}
		\toprule
           \toprule
           &BG&MN40(1k)&Segpart(C.mIoU)\\
\toprule    
           Transformer &95.01&93.92&84.27\\
         ~~$+$Linear Proj. &95.11&93.97&84.32\\
            ~~$+$Gate Fusion &95.35&94.01&84.64\\
         \textbf{$+$Full} &\textbf{95.53}&\textbf{94.16}&\textbf{84.88}\\
\toprule
        \toprule
	\end{tabular}
    \caption{Flow-Conditioned Fine-Tuning Architecture.}
\label{fine-tuning}
\end{table}
\textbf{Flow-Conditioned Fine-Tuning Architecture.} Table \ref{fine-tuning} presents the specific impact of the flow-conditioned enhancement on downstream performance. We first use a flow-conditioned projection layer to transform the pre-trained MFT’s geometric distribution into a feature-aligned space. Then, a learnable gating fusion mechanism adaptively controls the fusion strength, allowing the network to dynamically leverage pre-trained knowledge. While using projection or gating alone brings limited gains, their combination (FULL) boosts segmentation performance to 84.88\%, improving by 0.61\% over the baseline.

\begin{table}[t!]
\renewcommand\arraystretch{1.0}
    \small
	\begin{tabular}[width=0.70\linewidth]{p{2.5cm}p{1.1cm}<{\centering}p{1.6cm}<{\centering}p{1.5cm}<{\centering}}
		\toprule
           \toprule
           &OBJ\_BG&OBJ\_ONLY&PB\_T50\_RS\\
\toprule    
           Deterministic MAE &90.02&88.29&85.18\\
           +DDPM &93.29&91.91&87.61\\
           +Rectified Flow &94.84&92.60&89.60\\
           +MeanFlow(Ours) &\textbf{95.18}&\textbf{92.77}&\textbf{90.63}\\
\toprule
        \toprule
	\end{tabular}
    \caption{Performance comparison of different probabilistic modeling methods.}
    \label{method}
\end{table}
\textbf{Comparison of Probabilistic Modeling Methods.} To validate the superiority of MeanFlow over other probabilistic generative methods, multiple probabilistic modeling techniques are compared under the same experimental settings. Table \ref{method} presents the performance results of different methods. In terms of classification accuracy, MeanFlow achieves the highest accuracy, outperforming diffusion models and other flow matching variants. Regarding Numerical Stability, the time-interval MeanFlow effectively reduces gradient variance during pre-training, resulting in a more stable and reliable training process. Specifically, it satisfies the inequality:
\begin{equation}\mathrm{Var}[\mathbf{u}_{t,s}(\mathbf{z}_t)]\leq\frac{1}{(s-t)^2}\int_t^s\mathrm{Var}[\mathbf{v}_\tau(\mathbf{z}_\tau)]d\tau,\end{equation}
where $\mathbf{v}_\tau(\mathbf{z}_\tau)$ denotes the instantaneous velocity field at time $\tau$. A detailed proof is provided by \textbf{Theorem C} in the Supplementary Material.

\begin{table}[t!]
\renewcommand\arraystretch{1.0}
    \centering
    \small
	\begin{tabular}{
        >{\raggedright\arraybackslash}p{2cm} 
        >{\centering\arraybackslash}p{3.6cm}
        >{\centering\arraybackslash}p{1.4cm}
    }
		\toprule
                \toprule
        Loss& Sampling strategy & OBJ\_BG \\
        \midrule
        \multirow{2}{*}{MSE} 
            & Uniform Sampling & 95.01 \\
            & Lognormal Sampling & 94.84 \\
        \midrule
        \multirow{2}{*}{Adaptive L2} 
            & Uniform Sampling & \textbf{95.53} \\
            & Lognormal Sampling & 95.35 \\
        \toprule
                \toprule
	\end{tabular}
    \caption{Comparison of time sampling strategies and loss functions.}
    \label{time_sample}
\end{table}

\begin{figure}[t!]
  \centering
  \includegraphics[width=3.4in]{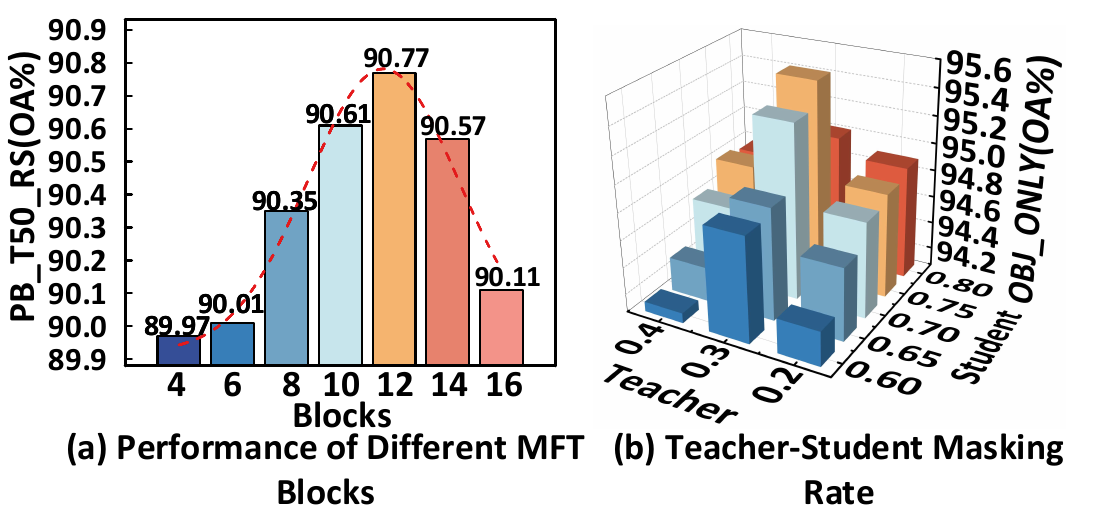}
 \caption{MFT and teacher-student hyperparameter analysis.}
  \label{fig5}
\end{figure}

\textbf{MFT Hyperparameter.} We systematically analyze key design choices of the MFT, including network block, time sampling strategy, and loss function. As shown in Figure \ref{fig5}(a), 12-layer MFT blocks achieve the best trade-off between performance and computational cost. Table \ref{time_sample} compares different time scheduling methods, where uniform sampling over the interval $[0,1]$ provides the most stable training signals for MeanFlow and outperforms log-normal sampling. Regarding the loss function, the adaptive L2 loss leverages a weighting mechanism to better handle sample-wise loss variations, offering more stable and efficient optimization compared to MSE loss.

\textbf{Mask Ratio Configuration.} Figure \ref{fig5}(b) shows the impact of various mask ratio settings on performance. The optimal configuration uses a 30\% mask ratio for the teacher and 75\% for the student, achieving the best results. This setup allows the teacher model to retain rich geometric details while the student focuses on learning higher-level semantic abstractions. An approximate mask ratio difference of 0.45 achieves the best balance; smaller differences fail to provide sufficient complementarity, while larger differences increase alignment difficulty and degrade knowledge transfer effectiveness.
\section{Related Work}
Point-MAE \cite{pang2022masked} and Point-M2AE \cite{zhang2022point} apply MAE to point cloud with fixed or multi-scale masking, but struggle with geometric diversity and point-wise reconstruction. MaskPoint \cite{liu2022masked} introduces geometry-aware masking, while PTM \cite{cheng2024ptm} and Point-FEMAE \cite{zha2024towards} propose density or a parallel masking strategy, though all involve complex tuning and limited semantic understanding. ReCon \cite{qi2023contrast} propose tri-modal contrastive learning frameworks to enhance semantic representation capabilities of 3D MAE. Flow Matching \cite{lipman2022flow} enables direct learning of probability flows, and MeanFlow \cite{geng2025mean} improves training stability via average velocity prediction. PointFM \cite{cheng2025pointfm} applies rectified flow matching to point cloud representation learning but requires careful design of flow predictors and suffers from training instability. Methods like REPA \cite{yu2024representation} demonstrates the benefits of self-distillation and cross-stage alignment. I-DAE \cite{chen2024deconstructing} demonstrates that hidden states of diffusion models can learn discriminative representations. Our work builds on these insights to propose a unified Dual Self-Representation Alignment mechanism across masking ratios and temporal states.
\section{Conclusion}
This paper proposes Point-SRA, a self-supervised representation learning framework. Firstly, it reveals for the complementary features of different mask ratios through systematic study. Secondly, we innovatively explore MeanFlow for 3D SSRL by probabilistic modeling and representation guidance. In addition, a dual self-representation alignment mechanism is designed by combining mask ratio alignment and MeanFlow temporal alignment. Finally, we build a dedicated fine-tuning network that leverages geometric distribution knowledge learned during pre-training via flow vector condition fusion. Extensive experiments validate the effectiveness of Point-SRA, achieving strong results on many standard benchmarks. Future work will focus on exploring efficient latent states knowledge distillation methods to further advance 3D SSRL.
\section{Acknowledgements}
This paper was funded in part by the Applied Technology Research and Development Project of Beilin District, Xi’an City (Grant No. GX2305), the Natural Science Basic Research Plan of Shaanxi Province, China (Grant No. 2025JC-JCQN-091), and the Technology Innovation Leading Program of Shaanxi Province (Program No. 2024QY-SZX-23).
\bibliography{aaai2026}
\clearpage
\newpage
\begin{appendix}
\setcounter{equation}{0}
\section{Supplementary Material}

\subsection{Details of Observations}

\begin{figure}[h!]
  \centering
  \includegraphics[width=3.3in]{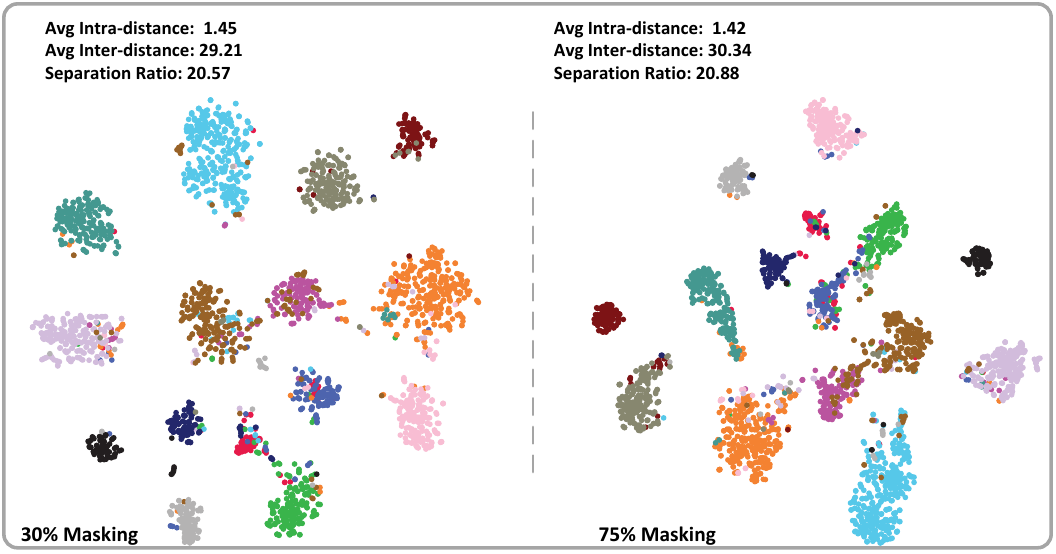}
 \caption{\textbf{Teacher (left) and student (right) t-SNE visualization}. t-SNE shows the teacher preserves geometry but blurs class boundaries (Inter-distance), while the student yields clearer separation and semantic clusters (Intra-distance).}
  \label{fig6}
\end{figure}

 As shown in Figure \ref{fig6}, low masking ratios ($\leq$30\%) emphasize reconstructing local structures like curvature and edges, whereas high ratios ($\geq$75\%) drive reasoning about global structure, enhancing semantic features. Model trained at low masking ratios excels in detail preservation and feature diversity, while high-rate model achieves stronger semantic abstraction and consistency, as confirmed by t-SNE visualizations.

\textbf{Details of Theorem A: Mask Ratio Complementarity.}

Given a point cloud $\mathcal{P}\in\mathcal{R}^{N\times3}$, it is partitioned into $G$ local regions $\mathcal{N}1,\mathcal{N}_2,...,\mathcal{N}_G$ via FPS. A subset of $\lfloor r\cdot G\rfloor$ regions is randomly masked, where $r\in[0,1]$ denotes the mask ratio. Let $\mathcal{V}$ denote the set of visible region indices, then the visible regions are $\mathcal{X}_r=\{\mathcal{N}_i:i\in\mathcal{V}\}$. The encoder $f_{\theta_r}:\mathcal{X}_r\rightarrow\mathcal{R}^d$ maps visible regions to a $d$-dimensional representation space.

\textbf{Theorem A: Masking Ratio Complementarity}. For masking ratios $r_l<r_h$, under the information bottleneck framework, the corresponding optimal encoders $f_{\theta_l^\ast}$ and $f_{\theta_h^\ast}$ satisfy:
\begin{equation}\mathcal{I}(\mathcal{P};f_{\theta_l^\ast}(\mathcal{X}_{r_l}))>\mathcal{I}(\mathcal{P};f_{\theta_h^\ast}(\mathcal{X}_{r_h}))\end{equation}
\begin{equation}\mathcal{C}(f_{\theta_h^\ast}(\mathcal{X}_{r_h}))>\mathcal{C}(f_{\theta_l^\ast}(\mathcal{X}_{r_l}))\end{equation}
where $\mathcal{I}(\cdot;\cdot)$ denotes mutual information and $\mathcal{C}(\cdot)$ denotes the semantic compression degree of the representation. 

\textbf{Proof}:

Based on the information bottleneck principle, the optimization objective for the low mask ratio encoder is:
\begin{equation}\min_{\theta_l} {E}_{\mathcal{P}} \left[ \mathcal{L}_{{recon}}(\mathcal{P}_{\mathcal{M}_{r_l}}, \hat{\mathcal{P}}_{\mathcal{M}_{r_l}}) \right] + \beta_l \mathcal{J}(\mathcal{X}_{r_l}; f_{\theta_l}(\mathcal{X}_{r_l}))\end{equation}
and for the high mask ratio encoder:
\begin{equation}\min_{\theta_h} {E}_{\mathcal{P}} \left[ \mathcal{L}_{{recon}}(\mathcal{P}_{\mathcal{M}_{r_h}}, \hat{\mathcal{P}}_{\mathcal{M}_{r_h}}) \right] + \beta_h \mathcal{J}(\mathcal{X}_{r_h}; f_{\theta_h}(\mathcal{X}_{r_h}))\end{equation}
Since $r_l<r_h$, it follows that $|\mathcal{X}_{r_l}|>|\mathcal{X}_{r_h}|,$ meaning the low mask ratio model receives more input information. Denote the entropy of the masked regions as $ \mathcal{H}(\mathcal{P}_\mathcal{M})$. Due to the randomness in masking, we have:
\begin{equation}{E}[\mathcal{H}(\mathcal{P}_{\mathcal{M}_{r_l}})] \approx \left. {E}[\mathcal{H}(\mathcal{P}_{\mathcal{M}_{r_h}})] \right|\end{equation}
which indicates that the reconstruction tasks under different mask ratios have similar intrinsic complexity. To achieve similar reconstruction performance:
\begin{equation}{E}[\mathcal{L}_{{recon}}^{(l)}] \approx {E}[\mathcal{L}_{{recon}}^{(h)}]\end{equation}
Since the low mask ratio model has access to more information, its information bottleneck regularization weight satisfies $\beta_l < \beta_h$. At optimality, the Lagrangian condition is:
\begin{equation}\frac{\partial\mathcal{L}_{{recon}}}{\partial f_\theta}+\beta\frac{\partial\mathcal{I}(\mathcal{X};f_\theta(\mathcal{X}))}{\partial f_\theta}=0\end{equation}

Because $\beta_l < \beta_h$ and reconstruction losses are similar, the representation of the low mask ratio model retains more geometric information about $\mathcal{P}$:
\begin{equation}\mathcal{I}\left(\mathcal{P};f_{\theta_l^\ast}\left(\mathcal{X}_{r_l}\right)\right)>\mathcal{I}\left(\mathcal{P};f_{\theta_h^\ast}\left(\mathcal{X}_{r_h}\right)\right)\end{equation}
Semantic compression is defined as:
\begin{equation}\mathcal{C}(f_\theta(\mathcal{X}))=\frac{\mathcal{H}(f_\theta(\mathcal{X}))}{\mathcal{I}(\mathcal{P}_{{semantic}};f_\theta(\mathcal{X}))}\end{equation}
where $\mathcal{P}_{\mathrm{semantic}}$ denotes semantic information of the point cloud. Due to stronger compression constraints, the high mask ratio model’s representation tends to retain semantic information while discarding geometric details:
\begin{equation}\mathcal{C}\left(f_{\theta_h^\ast}\left(\mathcal{X}_{r_h}\right)\right)>\mathcal{C}\left(f_{\theta_l^\ast}\left(\mathcal{X}_{r_l}\right)\right)\end{equation}

\textbf{Corollary: Representation Complementarity.}
There exist projection functions $\pi_{{geo}}:\mathcal{R}^d\rightarrow\mathcal{R}^{d_{{geo}}}$ and $\pi_{{sem}}:\mathcal{R}^d\rightarrow\mathcal{R}^{d_{{sem}}}$ such that:
\begin{align} 
    \| \pi_{{geo}}(f_{\theta_l^*}(\mathcal{X}_{rl})) & - \pi_{{geo}}(\mathcal{P}) \|_F \nonumber \\
    & < \| \pi_{{geo}}(f_{\theta_h^*}(\mathcal{X}_{rh})) - \pi_{{geo}}(\mathcal{P}) \|_F 
\end{align}
\begin{align}
    \| \pi_{{sem}}(f_{\theta_h^*}(\mathcal{X}_{rh})) & - \pi_{{sem}}(\mathcal{P}) \|_F \nonumber \\
    & < \| \pi_{{sem}}\! (f_{\theta_l^*}(\mathcal{X}_{rl})) - \pi_{{sem}}(\mathcal{P}) \|_F
\end{align} 

\textbf{Proof}:

We define the geometric projection function as:
\begin{equation}\pi_{{geo}}(z) = \arg\min_{g} {E}_{\mathcal{P}} \left[ \| {Geo}(\mathcal{P}) - g \|^2 \mid f(\mathcal{X}) = z \right]\end{equation}
\begin{equation}\pi_{{sem}}(z) = \arg\min_{s} \mathbb{E}_{\mathcal{P}} \left[ \| {Sem}(\mathcal{P}) - s \|^2 \mid f(\mathcal{X}) = z \right]\end{equation}
According to \textbf{Theorem A}, the low mask ratio encoder $f_{\theta_l^\ast}$ preserves more original information, and thus exhibits a stronger correlation with geometric features:
\begin{align} \label{eq:geo_correlation}
    {Corr} \bigl( & \! \pi_{{geo}}\bigl(f_{\theta_l^*}(\mathcal{X}_{r_l})\bigr), {Geo}(\mathcal{P}) \bigr) \nonumber \\
    & > {Corr} \bigl( \! \pi_{{geo}}\bigl(f_{\theta_h^*}(\mathcal{X}_{r_h})\bigr), {Geo}(\mathcal{P}) \bigr)
\end{align}
This directly implies:
\begin{align} \label{eq:geo_norm_inequality}
    \bigl\lVert \pi_{{geo}} & \! \bigl(f_{\theta_l^*}(\mathcal{X}_{r_l})\bigr) - {Geo}(\mathcal{P}) \bigr\rVert_F \nonumber \\
    & < \bigl\lVert \pi_{{geo}} \! \bigl(f_{\theta_h^*}(\mathcal{X}_{r_h})\bigr) - {Geo}(\mathcal{P}) \bigr\rVert_F
\end{align}
On the other hand, since the high mask ratio encoder $f_{\theta_h^\ast}$ faces stronger compression constraints, its representation is more semantically biased:
\begin{align} \label{eq:sem_correlation}
    {Corr} \bigl( & \! \pi_{{sem}}\bigl(f_{\theta_h^*}(\mathcal{X}_{r_h})\bigr), {Sem}(\mathcal{P}) \bigr) \nonumber \\
    & < {Corr} \bigl( \! \pi_{{sem}}\bigl(f_{\theta_l^*}(\mathcal{X}_{r_l})\bigr), {Sem}(\mathcal{P}) \bigr)
\end{align}

Therefore:
\begin{align}
    \| \pi_{{sem}} & \! (f_{\theta_h^*}(\mathcal{X}_{rh})) - \pi_{{sem}}(\mathcal{P}) \|_F \nonumber \\
    & < \| \pi_{{sem}}\! (f_{\theta_l^*}(\mathcal{X}_{rl})) - \pi_{{sem}}(\mathcal{P}) \|_F
\end{align} 

\textbf{Details of Theorem B: Reconstruction Uncertainty.}

When the mask ratio is within a reasonable range, the entropy of the reconstruction distribution satisfies:
\begin{equation}H(p(\mathcal{P}_\mathcal{M}|\mathcal{P}_\mathcal{V}))>0\end{equation}

\textbf{Proof:}

Given the visible region $\mathcal{P}_\mathcal{V}$, the masked region $\mathcal{P}_\mathcal{M}$ follows a conditional distribution:
\begin{equation}p(\mathcal{P}_\mathcal{M}|\mathcal{P}_\mathcal{V})=\int_{\Theta}p(\mathcal{P}_\mathcal{M}|\mathcal{P}_\mathcal{V},\omega)p(\omega|\mathcal{P}_\mathcal{V})d\omega\end{equation}
This distribution is governed by several constraints, including surface continuity, which refers to the smoothness of the underlying geometry; topological consistency, meaning the preservation of genus and connectivity; and physical plausibility, which entails a reasonable shape, volume, and mass distribution.

\begin{figure}[htbp!]
  \centering
  \includegraphics[width=3.3in]{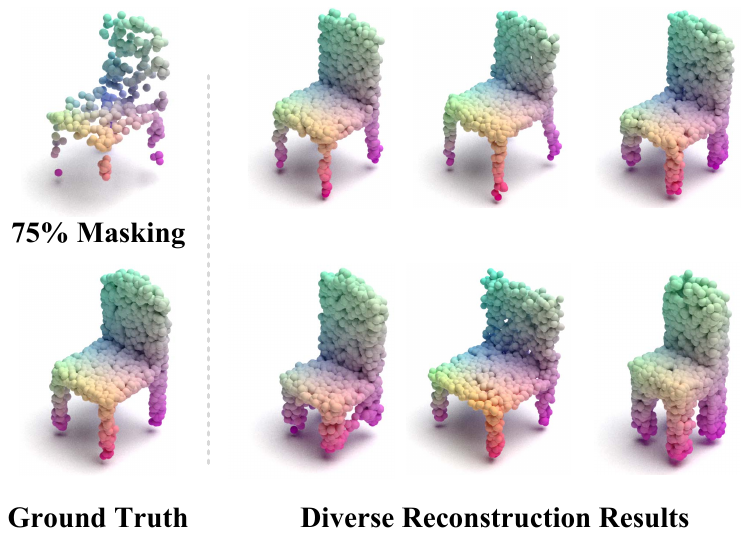}
 \caption{Diverse reconstruction results given partial observation.}
  \label{fig7}
\end{figure}

Under these constraints, multiple plausible completions exist for a given partial observation. As shown in Figure \ref{fig7}, a chair with a known seat and backrest may have several valid leg structures (straight, curved, etc.).

Let $\Omega={\mathcal{P}_\mathcal{M}^{(1)},\mathcal{P}_\mathcal{M}^{(2)},...,\mathcal{P}_\mathcal{M}^{(K)}}$, with $K\geq2$, be the set of feasible completions, each with non-zero probability:
\begin{equation}p(\mathcal{P}_\mathcal{M}^{(i)}|\mathcal{P}_\mathcal{V})>\epsilon>0,\forall i\in{1,2,...,K}\end{equation}
Then the entropy is strictly positive:
\begin{equation}H(p(\mathcal{P}_\mathcal{M}|\mathcal{P}_\mathcal{V}))\geq-\sum_{i=1}^{K}\ p(\mathcal{P}_\mathcal{M}^{(i)}|\mathcal{P}_\mathcal{V})\log\ p(\mathcal{P}_\mathcal{M}^{(i)}|\mathcal{P}_\mathcal{V})>0\end{equation}
The inequality holds because $K\geq2$ and the distribution is non-degenerate., there does not exist a unique “correct” reconstruction.

\textbf{Theorem C: Numerical Stability of MeanFlow.}

Compared to instantaneous flow, MeanFlow provides better numerical stability. Specifically, the variance of the MeanFlow satisfies:
\begin{equation}\mathrm{Var}[\mathbf{u}_{t,s}(\mathbf{z}_t)]\leq\frac{1}{(s-t)^2}\int_t^s\mathrm{Var}[\mathbf{v}_\tau(\mathbf{z}_\tau)]d\tau\end{equation}

\textbf{Proof:}

The MeanFlow vector is defined as:
\begin{equation}\mathbf{u}_{t,s}(\mathbf{z}_t)=\frac{1}{s-t}\int_{t}^{s}\mathbf{v}_\tau(\mathbf{z}_\tau)d\tau\end{equation}
By the variance properties of integrals:
\begin{equation}\mathrm{Var}[\mathbf{u}_{t,s}]=\frac{1}{(s-t)^2}\mathrm{Var}\left[\int_{t}^{s}\mathbf{v}_{\tau}(\mathbf{z}_{\tau})\mathrm{d}\tau\right]\end{equation}
Using the definition of variance of an integral:
\begin{equation}\mathrm{Var}\left[\int_t^s\mathbf{v}_\tau(\mathbf{z}_\tau)d\tau\right]=\int_t^s\int_t^s\mathrm{Cov}[\mathbf{v}_{\tau_1}(\mathbf{z}_{\tau_1}),\mathbf{v}_{\tau_2}(\mathbf{z}_{\tau_2})]d\tau_1d\tau_2\end{equation}
Assuming $\mathbf{v}_\tau$ follows a Markov process with exponential decay of covariance:
\begin{equation}\operatorname{Cov}[\mathbf{v}_{\tau_1}(\mathbf{z}_{\tau_1}),\mathbf{v}_{\tau_2}(\mathbf{z}_{\tau_2})]\leq C\cdot e^{-\lambda|\tau_1-\tau_2|}\end{equation}
Then the double integral admits an upper bound:
\begin{equation}\begin{aligned}\int_{t}^{s}\int_{t}^{s}e^{-\lambda|\tau_1-\tau_2|}d\tau_1d\tau_2 & =2\int_{t}^{s}\int_{\tau_2}^{s}e^{-\lambda(\tau_1-\tau_2)}d\tau_1d\tau_2\\  & =\frac{2}{\lambda}\int_{t}^{s}{(1-e^{-\lambda(s-\tau_2)})d\tau_2}^{}\\  & \leq\frac{2}{\lambda}(s-t)\end{aligned}\end{equation}
Therefore,
\begin{equation}\mathrm{Var}[\int_t^s\mathbf{v}_\tau(\mathbf{z}_\tau)d\tau]\leq\frac{2}{\lambda}(s-t)\mathrm{~}\Rightarrow\mathrm{Var}[\mathbf{u}_{t,s}]\leq\frac{2\mathcal{C}}{\lambda(s-t)}\end{equation}
This confirms that MeanFlow has lower variance than the instantaneous flow and is thus more stable. Alternatively, using Cauchy-Schwarz:
\begin{equation}\operatorname{Cov}[\mathbf{v}_{\tau_1},\mathbf{v}_{\tau_2}]\leq\sqrt{\operatorname{Var}[\mathbf{v}_{\tau_1}]\cdot\operatorname{Var}[\mathbf{v}_{\tau_2}]}\end{equation}
Let $g(\tau)=\sqrt{\mathrm{Var}[\mathbf{v}_\tau]}$, then:
\begin{equation}\begin{aligned}
\mathrm{Var}\left[\int_t^s\mathbf{v}_\tau(\mathbf{z}_\tau)d\tau\right] & \leq\left(\int_t^sg(\tau)d\tau\right)^2 \\
 & \leq(s-t)\int_t^s\mathrm{Var}[\mathbf{v}_\tau(\mathbf{z}_\tau)]d\tau
\end{aligned}\end{equation}
So finally:
\begin{equation}\begin{aligned}
\mathrm{Var}[\mathbf{u}_{t,s}]\leq\frac{1}{(s-t)^2}\cdot(s-t)\int_{t}^s\mathrm{Var}[\mathbf{v}_\tau(\mathbf{z}_\tau)]d\tau \\
=\frac{1}{s-t}\int_{t}^s\mathrm{Var}[\mathbf{v}_\tau(\mathbf{z}_\tau)]d\tau
\end{aligned}\end{equation}

\subsection{Experimental Details}
Point-SRA is pre-trained on the ShapeNet \cite{chang2015shapenet}, following the same setting as previous SSRL \cite{pang2022masked} methods to enable fair comparisons. ShapeNet consists of 55 object categories with approximately 50K CAD models. The input is downsampled to 1024 points and partitioned into 64 local regions, each containing 32 points. We apply FPS to select region centers, which are fed into the MFT. For multi-modal condition, a pre-trained Vision Transformer \cite{dosovitskiy2020image} is used to encode 2D rendered views of point cloud, while CLIP \cite{radford2021learning} is employed to encode corresponding text descriptions. Both the image and text encoders are frozen during pre-training. Point-SRA is optimized using AdamW with an initial learning rate of 5e-4, a weight decay of 5e-2, and cosine learning rate scheduling over 300 epochs. The batch size is set to 128. The Transformer encoder and decoder both operate on 384-dimensional features with 6 attention heads. The encoder contains 12 layers, while the decoder comprises 4 layers. For downstream fine-tuning, we adopt the frozen flow-conditioned architecture. The input is a complete point cloud downsampled to 2048 points and grouped into 128 local regions of 32 points each. On the more challenging ScanObjectNN benchmark, including its three variants (OBJ\_BG, OBJ\_ONLY, and PB\_T50\_RS), we use AdamW with an initial learning rate of 2e-5, a weight decay of 5e-2, and cosine annealing over 500 epochs. The batch size is set to 32. For ModelNet40, the input point cloud is sampled to 1024 (1k) and 8192 (8k) points, with other configurations consistent with ScanObjectNN. The classification head follows the same architecture as Point-MAE. For 3D detection, we use the AdamW optimizer with an initial learning rate of 1e-4 and a weight decay of 0.1. The model is trained with a batch size of 8 using a cosine learning rate schedule. 

\subsection{Additional Experiments}

\begin{table}[t!]
    \centering
    \footnotesize
 \renewcommand\arraystretch{1.3}
\begin{tabular}[width=0.80\linewidth]{p{2.9cm} p{2cm}<{\centering}p{2cm}<{\centering}}
    \toprule
    \toprule
    \multirow{2}{*}{\textbf{Method}} & \multicolumn{2}{c}{\textbf{ModelNet}}\\
    \cmidrule(lr){2-3}
     & 1k\_P & 8k\_P\\
    \midrule
        \multicolumn{3}{c}{\textit{Supervised Learning Only}} \\
        \midrule
        PointNet~\shortcite{qi2017pointnet} & 89.2 & 90.8 \\
        PointNet++~\shortcite{qi2017pointnet++} & 90.7 & 91.9  \\
        PointNeXt~\shortcite{qian2022pointnext} & 94.0 & -- \\
        P2P~\shortcite{wang2022p2p}  & 94.0 & -- \\
        \midrule
        \multicolumn{3}{c}{\textit{Single-Modal Self-Supervised Representation Learning}} \\
        \midrule
        Point-BERT~\shortcite{yu2022point} & 93.2 & 93.8 \\
        Point-MAE~\shortcite{pang2022masked} & 93.8 & 94.0 \\
        Point-M2AE~\shortcite{zhang2022point}   & 94.0 & --  \\
        Point-FEMAE~\shortcite{zha2024towards}  & 94.0 & -- \\        
        Point-JEPA~\shortcite{saito2025point}& 93.8 & -- \\
       \midrule
        \multicolumn{3}{c}{\textit{Cross-Modal Self-Supervised Representation Learning}} \\
        \midrule
        ACT~\shortcite{dong2022autoencoders}   & 93.7 & 94.0\\
        I2P-MAE\shortcite{zhang2023learning} & 93.7 & --  \\
        ReCon~\shortcite{qi2023contrast} & 94.1 & 94.3  \\
        \textbf{Point-SRA (Ours)}  & \textbf{94.3} & \textbf{94.5} \\
        \bottomrule
    \bottomrule
    \end{tabular}
    \caption{Classification accuracy (\%) on the ModelNet40. We report the performance on 1k and 8k point settings for ModelNet40.}
    \label{table_MD}
\end{table} 

\begin{table}[t!]
\renewcommand\arraystretch{1.2}
    \footnotesize
	\begin{tabular}[width=0.80\linewidth]{p{2.8cm} p{2cm}<{\centering}p{1cm}<{\centering}p{1cm}<{\centering}}
		\toprule
           \toprule
         Method & Journal or Conference& Cls. mIoU(\%)&Ins. mIoU(\%)\\
\toprule
      PointNet\shortcite{qi2017pointnet}&CVPR&	80.4&	83.7\\
      PointNet++\shortcite{qi2017pointnet++}&NeurIPS&81.9&85.1\\
      DGCNN\shortcite{wang2019dynamic}&TOG&82.3&85.2\\
\toprule
    \multicolumn{4}{c}{\textit{Self-Supervised Representation Learning}}\\
    \toprule
      Point-BERT\shortcite{yu2022point}&CVPR&84.1&85.6\\
      Point-MAE\shortcite{pang2022masked}&ECCV&	-&86.1\\
      PointMamba\shortcite{liang2024pointmamba}&NeurIPS&84.3&	86.2\\
\toprule
        \multicolumn{4}{c}{\textit{Cross-Modal Self-Supervised Representation Learning}}\\
        \toprule
      ACT\shortcite{dong2022autoencoders}&ICLR&84.7&86.1\\
      ReCon\shortcite{qi2023contrast}&ICML&84.8&86.4\\
      \textbf{Point-SRA(Ours)}&-&	\textbf{84.9}&	\textbf{86.7}\\
\toprule
        \toprule
	\end{tabular}
    \caption{ Part segmentation results (mIoU\%) of various methods on ShapeNetPart: Cls. mIoU (\%) and Ins. mIoU (\%) denote mean IoU across all part categories and instances.}
\label{partseg} 
\end{table}

\textbf{Synthetic Object Classification.} We evaluate the pretrained model on synthetic object classification benchmark. ModelNet40 \cite{wu20153d} consists of 12,311 synthetic CAD models spanning 40 common object categories. As shown in Table \ref{table_MD}, our Point-SRA reaches 94.3 with 1k points and 94.5\% with 8k points, demonstrating competitive performance. 

\textbf{Part Segmentation.} In the ShapeNetPart \cite{chang2015shapenet} benchmark, Point-SRA is challenged to recognize and segment fine-grained semantic parts of objects. ShapeNetPart contains 16,881 3D shapes from 16 categories, with each shape annotated with 2 to 6 functional parts, totaling 50 unique part classes. We adopt the standard metrics of category mean IoU (Cls.mIoU) and instance mean IoU (Ins.mIoU), where the former measures the average segmentation quality per part category, and the latter evaluates the overall segmentation quality per shape instance. As shown in Table \ref{partseg}, Point-SRA achieves 84.9\% in C.mIoU and 86.7\% in I.mIoU, surpassing Point-MAE by 0.7 and 0.6 respectively, and outperforming ReCon by 0.1 and 0.3.

\subsection{Additional Ablation Study}

\textbf{Cross-Modal Condition Fusion Strategies.} We further analyze the effectiveness of different fusion strategies. As shown in Table \ref{Cross-Modal}, the simple addition fusion strategy achieves the best performance while being computationally efficient and easy to implement. This approach avoids complex cross-modal attention mechanisms and heavy computation by linearly projecting different modal features into a unified space and directly summing them. It preserves the independence of each modality’s information while enabling effective integration.

\textbf{EMA Parameter Analysis.} We analyze the impact of the momentum parameter in the EMA update mechanism on the performance of the teacher-student framework. Table \ref{m} presents the comparison results for different momentum coefficients. Experimental results show that m=0.996 is the optimal choice, achieving the highest classification accuracy among all tested values.

\begin{table}[t!]
    \centering
    \begin{minipage}[t]{0.2\textwidth} 
        \centering
        \footnotesize
        \renewcommand\arraystretch{1.25}
        \captionsetup{justification=centering, singlelinecheck=false}          
        \begin{tabular}{p{2.1cm}<{\centering}p{1.6cm}<{\centering}}
            \toprule
            \toprule
            Method & OBJ\_ONLY \\
            \midrule
            Concatenate & 92.25 \\
            Cross Attention & 92.60 \\
            \textbf{Add} & \textbf{92.77} \\
            \bottomrule
            \bottomrule
        \end{tabular}
        \caption{Comparison of cross-modal fusion strategies.}
        \label{Cross-Modal}
    \end{minipage}
    \hfill 
    \begin{minipage}[t]{0.2\textwidth}
        \centering
        \footnotesize
           \captionsetup{justification=centering, singlelinecheck=false} 
        \begin{tabular}{p{1.0cm}<{\centering}p{1.9cm}<{\centering}}
            \toprule
            \toprule
            $m$ & PB\_T50\_RS \\
            \midrule
            0.99 & 90.39 \\
            0.995 & 90.57 \\
            \textbf{0.996} & \textbf{90.77} \\
            0.997 & 90.71 \\
            \bottomrule
            \bottomrule
        \end{tabular}
        \caption{Sensitivity analysis of EMA momentum parameter.}
            \label{m}
    \end{minipage}

\end{table}

\textbf{Stop-gradient Effectiveness Analysis.} We compare the performance differences with and without applying stop-gradient. As shown in Table \ref{stop}, without stop-gradient, performance degrades significantly: ModelNet40 accuracy drops to 93.4\% and ScanObjectNN accuracy to 90.38\%. This is because mutual gradient dependencies between student and teacher models can cause representation collapse and training divergence. Applying full stop-gradient achieves the best results, with ModelNet40 accuracy reaching 94.2\% and ScanObjectNN accuracy 90.77\%. The stop-gradient mechanism blocks gradient backpropagation to the teacher model, ensuring its parameters update solely via the EMA mechanism, thereby maintaining the unidirectionality and stability of the knowledge distillation process.

\begin{table}[t!]
\renewcommand\arraystretch{1.2}
    \footnotesize
	\begin{tabular}[width=0.80\linewidth]{p{2.1cm}<{\centering} p{2.5cm}<{\centering}<{\centering}p{2.5cm}<{\centering}}
		\toprule
           \toprule
          Stop-grad&PB\_T50\_RS&ModelNet40(1k\_P)\\
\toprule
           \ding{55}&90.38&93.4\\
           \checkmark&\textbf{90.77}&\textbf{94.2}\\   
\toprule
        \toprule
	\end{tabular}
    \caption{Stop-gradient effectiveness analysis.}
\label{stop}
\end{table}

\textbf{Loss Weight Analysis.} We tune the loss weights in Point-SRA to find the best balance. Table \ref{weight} shows that $\lambda_{flow}$=0.5, $\lambda_{mae-sra}=0.2$, and $\lambda_{mft-sra}=0.2$ yield optimal results. A low flow loss weight (e.g., 0.3) undermines the effectiveness of probabilistic modeling, whereas an excessively high weight (e.g., 0.7) disrupts the reconstruction objective by overemphasizing distribution learning. Both MAE and MFT alignment weights perform best at 0.2, highlighting the importance of balanced SRA. This setup effectively balances reconstruction, probabilistic modeling, and knowledge distillation.

\textbf{Cross-modal Condition Effectiveness.} As shown in Table \ref{modal}. Using only time embedding as a condition results in random point cloud structures and weak semantic consistency due to lack of semantic guidance. Adding image features significantly improves performance by providing rich visual-geometric priors. Text conditions bring further gains by offering high-level semantic cues, especially for object function and category understanding. Combining both modalities yields the best results, enabling flow matching to achieve precise geometry and accurate semantics.
\begin{table}[t!]
\renewcommand\arraystretch{1.2}
	\begin{tabular}[width=0.80\linewidth]{p{1.5cm}<{\centering}p{1.5cm}<{\centering}p{1.5cm}<{\centering}p{2.0cm}<{\centering}}
		\toprule
           \toprule
          $\lambda_{flow}$&$\lambda_{mae-sra}$&$\lambda_{mft-sra}$&OBJ\_ONLY\\
\toprule    
            \textbf{0.5}&\textbf{0.2}&\textbf{0.2}&\textbf{93.31}\\
            0.3&0.2&0.2&92.80\\
            0.7&0.2&0.2&93.11\\
            0.5&0.1&0.2&92.97\\
            0.5&0.2&0.1&93.28\\ 
            0.5&0.2&0.2&93.31\\
            0.5&0.2&0.3&92.84\\ 
            0.5&0.3&0.2&92.98\\
\toprule
        \toprule
	\end{tabular}
    \caption{Loss weight analysis.}
\label{weight}
\end{table}

\begin{table}[t!]
\renewcommand\arraystretch{1.2}
    \begin{tabular}{>{\centering\arraybackslash}m{2.2cm} 
                >{\centering\arraybackslash}m{2.2cm} 
                >{\centering\arraybackslash}m{2.6cm}} 
		\toprule
           \toprule
           Image&Text&PB\_T50\_RS(Acc\%)\\
\toprule    
            \ding{55}&\ding{55}&89.45\\
            \ding{55}&\checkmark&89.97\\
            \checkmark&\ding{55}&90.15\\
            \checkmark&\checkmark&\textbf{90.63}\\
\toprule
        \toprule
	\end{tabular}
    \caption{Impact of cross-modal conditions on MeanFlow performance.}
\label{modal}
\end{table}

\end{appendix}
\end{document}